\pdfoutput=1

\documentclass[11pt]{article}

\usepackage[preprint]{acl}

\usepackage{times}
\usepackage{latexsym}

\usepackage[T1]{fontenc}

\usepackage[utf8]{inputenc}

\usepackage{microtype}

\usepackage{inconsolata}

\usepackage{graphicx}

\usepackage{tabularx}
\usepackage{booktabs}
\usepackage{booktabs, multirow, multicol}
\usepackage{amsmath}
\usepackage{mathtools}
\usepackage{listings}
\usepackage{xcolor}

\definecolor{bg}{gray}{0.9}
\definecolor{c1}{RGB}{100,143,255}
\definecolor{c2}{RGB}{120,94,240}
\definecolor{c3}{RGB}{220,38,127}
\definecolor{c4}{RGB}{254,97,0}
\definecolor{c5}{RGB}{255,176,0}

\usepackage{enumitem}

\renewcommand{\arraystretch}{1.5}
\usepackage{array}
\usepackage{ragged2e}
\newcolumntype{P}[1]{>{\RaggedRight\hspace{0pt}}p{#1}}
\newcommand{\tabitem}{~\llap{-}~~}

\newcommand\emptyfootnote[1]{%
  \begingroup
  \renewcommand\thefootnote{}\footnote{#1}%
  \addtocounter{footnote}{-1}%
  \endgroup
}

\usepackage{listliketab}
\usepackage{todonotes}

\usepackage[framemethod=TikZ]{mdframed}

\usepackage[toc,page,header]{appendix}
\usepackage{minitoc}

\mdfdefinestyle{codestyle}{%
backgroundcolor=bg,
linecolor=black,
linewidth=1pt,%
rightline=false,leftline=false
}

\usepackage{tcolorbox}

\usepackage{amssymb}
\usepackage{pifont}
\usepackage{adjustbox}
\usepackage{fdsymbol}

\title{Tools Fail: Detecting Silent Errors in Faulty Tools}

\author{
  Jimin Sun\textsuperscript{1*} \hspace{2em} So Yeon Min\textsuperscript{2} \hspace{2em} Yingshan Chang\textsuperscript{2} \hspace{2em} Yonatan Bisk\textsuperscript{2}\\
  \textsuperscript{1}CohereAI \quad \textsuperscript{2}Carnegie Mellon University\\
  \texttt{jimin@cohere.com} \\}
  
\begin{document}
\maketitle
\begin{abstract}

Tools have become a mainstay of LLMs, allowing them to retrieve knowledge not in their weights, to perform tasks on the web, and even to control robots.  However, most ontologies and surveys of tool-use have assumed the core challenge for LLMs is choosing the tool. 
Instead, we introduce a framework for tools more broadly which guides us to explore a model's ability to detect ``silent'' tool errors, and reflect on how to plan. This more directly aligns with the increasingly popular use of models as tools.
We provide an initial approach to failure recovery with promising results both on a controlled calculator setting and embodied agent planning.
\end{abstract}
\section{Introduction}
Tools offer a convenient way to augment capabilities beyond text-based reasoning, 
from executing code to incorporating recent data through web search, and even facilitating multimodal interactions. 
While the term ``tool'' is often interpreted to mean offloading specific deterministic functions to external APIs, as 
tasks grow more complex, the definition is expanding 
to include learned modules such as translators and object detectors, as well as heuristics-based policies like search algorithms and robotic skills. LLMs themselves are also being used as tools, particularly as task planners in robotics, chained with vision models and robot policies to perform navigation and manipulation~\citep{saycanAhn2022DoAI,Huang2022LanguageMA,inner-monologue2022Huang,code-as-policies2022Liang,progprompt2022Singh,interactive-task-planning2023Li,xu2023creative,zeng2023socratic}. \emptyfootnote{*Work done while at Carnegie Mellon University.}

As tools take on more responsibilities, assessing and ensuring their reliability becomes crucial; a failure in one tool can trigger a cascade of errors, leading to complete task failure. 
Recent studies have suggested recovery mechanisms, such as correcting inputs based on API error messages \citep{pan2023logiclm, zhang2023selfedit, chen2023selfdebug, pan2023selfcorrectionsurvey}. However, most methods rely on two underlying assumptions: that accurate inputs guarantee flawless outputs, and that errors are accompanied by explicit signals. Yet, real-world scenarios challenge the premises, as failures often arise from unpredictable environmental dynamics and inherent inaccuracies of tools themselves.
\begin{figure}[t]
    \centering
    \includegraphics[width=\linewidth]{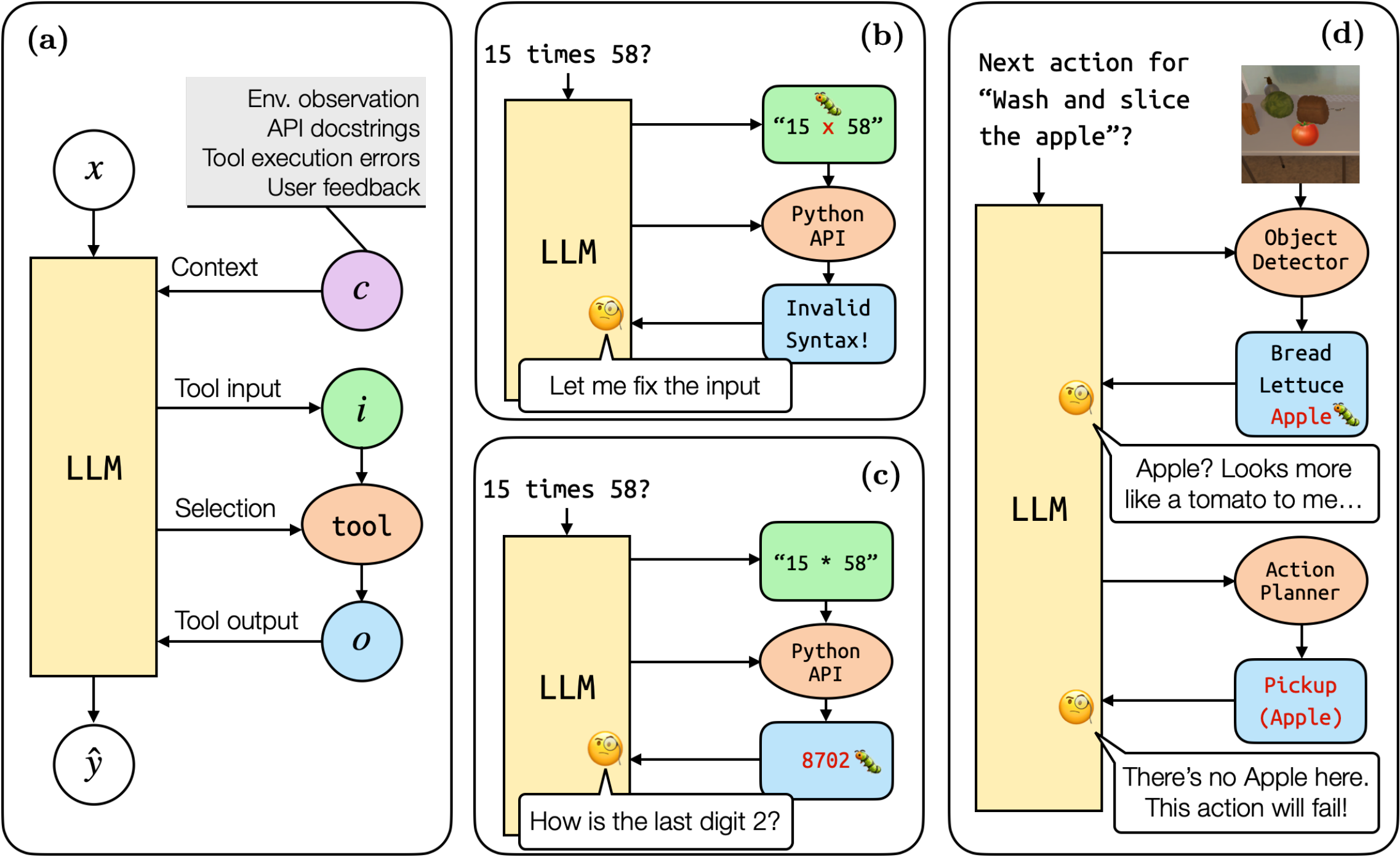}
    \captionsetup{font=footnotesize}
    \caption{\textbf{(a) Tool-use Overview}: Starting from an input $x$, the LLM generates inputs $i$ for the selected tool, and incorporates tool outputs $o$ to predict the final task output $\Hat{y}$. The context $c$ is used throughout the task. \textbf{(b) Correct Calculator} Incorrect tool inputs from the LLM leads to tool failure. The error messages can be leveraged for correction (Refine). \textbf{(c) Broken Calculator} Tool inputs are correct, but the tool itself silently produces false outputs. \textbf{(d) ALFRED} The first tool, Object Detector, misidentifies the Tomato in the image as an Apple, leading to error cascades in the next tool, the Action Planner.}
    \label{fig:overview_tooluse}
    \vspace{-10pt}
\end{figure}

This paper introduces a taxonomy to categorize sources of tool-related errors and recovery methods. We shed light on the often overlooked case: ``tool-based'' failures. As opposed to input-based errors which are often accompanied by error messages, most tool failures are ``silent.'' This poses unique reasoning challenges for the LLM, which must actively 1. detect the failure, 2. infer the source, and 3. plan recovery strategies. In this paper, we focus on the first step, detection, as it is the prerequisite for downstream fault assignment and recovery.

We investigate tool errors in two distinct settings (Fig.~\ref{fig:overview_tooluse}) -- a controlled environment where an LLM solves arithmetic problems using a broken calculator, and a more natural ``broken'' tool setting involving a multimodal instruction-following agent. 
We investigate whether LLMs can detect incorrect tool outputs without explicit error signals, to observe overtrusting of tools. Motivated by how humans detect tool failures based on internal expectations of correct outputs, we devise three in-context interventions, and find that LLMs \textit{can} learn to doubt tools and detect mistakes. 
Following the taxonomy, we further examine how much and what type of deviation is necessary to trigger the LLM's recognition of the tool error in each setting.

\section{Related Work}
\paragraph{Tools}
Text-based tools help compensate for LLMs' relative weakness in world knowledge and computational precision~\citep{rag-LewisPPPKGKLYR020,parisi2022talm,Gao2023PALPL,schick2023toolformer,yao2023react}. 
Multimodal tools allow LLMs to receive inputs from other modalities and generate grounded answers~\citep{Gupta_2023_CVPR_visual_prog,wu2023visual-chatgpt,yang2023mmreact,zeng2023socratic}. 
Outputs of Vision-Language models~\citep{CLIPRadford2021LearningTV}, Object Detectors, OCR models, and speech-to-text APIs~\citep{zeng2023socratic} have been added to the LLM's prompt, enabling zero-shot inference on multimodal tasks.

\paragraph{Agents} Research on LLM agents spans multi-step tasks in gaming~\citep{wang2023voyager, wu2024spring}, web navigation~\citep{qin2023toolllm, Shinn2023ReflexionLA, yao2023react}, and code generation~\citep{Shinn2023ReflexionLA, yao2023react}.
Most focus on the selection and utilization of tools~\citep{wang2023voyager, qin2023toolllm, wu2024spring}, and enhancing reasoning through self-evaluation and feedback~\citep{Shinn2023ReflexionLA, wang2023voyager, chen2023agentverse,xu2023creative, madaan2024self}. 

\paragraph{Adapting LLMs to tool-use} Existing works  use in-context learning (ICL)~\cite{lu2023chameleon, shen2024hugginggpt}, finetuning~\cite{schick2023toolformer}, and trial-and-error~\cite{wang2024llms} to adapt LLM to tool-use. However, the focus has been on adapting to ``newer'' tools, from demonstrations or documentations, and the question of tool reliability and recovering from ``unreliable'' tools has not been actively investigated. While malfunctioning APIs are preemptively filtered out in API-centric environments~\cite{qin2023toolllm}, the strategies for addressing ineffective learned tools, as in games~\cite{wang2023voyager, wu2024spring} or multimodal tasks \citep{zeng2022socratic}, have been less explored. Overall, existing approaches tend to amalgamate various tool failure modes under the umbrella term ``reasoning,'' focusing primarily on the most salient aspect of failure within their specific domain. In contrast, we distinctly identify and thoroughly analyze errors related to tool arguments, the tools themselves, and the alignment with environmental dynamics.

\section{Background}

\paragraph{Notation}
We outline a typical tool-use scenario in Fig.~\ref{fig:overview_tooluse}a with the following notation:  
\looseness=-1
\begin{align*}
    x &: \text{task input}   & i &: \text{tool input}\\
    \hat{y}& : \text{predicted task output}   &o & : \text{tool output}\\
    c &: \text{context information}    & t_\theta &: \text{tool} \\
\end{align*}
\looseness=-1
\vspace{-2.5em}

The LLM first selects tools and constructs tool-specific arguments $i$ from the task input $x$. Based on the tool result $o$, the final task prediction $\Hat{y}$ is made. Notably, the flexibility of LLMs as an interface allows inputs to be enriched with context information $c$ throughout the task. $c$ may include task specifics, API docstrings, any external feedback like error messages, or even previous action trajectories in interactive tasks. 

Additionally, we denote the oracle values of the input, output, context as $i^*$, $o^*$, and $c^*$. The tool input $i$ and output $o$ may contain inaccuracies since they are essentially outputs of preceding LLM/tool calls. Fig.~\ref{fig:overview_tooluse}b demonstrates a scenario where $i$ contains a mistake (\texttt{15 x 58} should be \texttt{15 * 58}). The context $c$ can also be incomprehensive or noisy, as they are approximations of the real world. 
Moreover, the tool $t_\theta$ can be suboptimal in multiple dimensions. For deterministic APIs, a suboptimal tool may have been chosen by an LLM~\citep{schick2023toolformer}.
For learned tools, the tool itself is an inherently imperfect parameterized model, thus $t_\theta$. 

\paragraph{Defining Error}
\label{sec:understanding_errors}
The suboptimality of $i$, $c$, and $t_\theta$ manifest as suboptimal tool outputs $o$, that deviate from $o^*$. The deviation can be as critical and explicit, leading to error messages in Fig.~\ref{fig:overview_tooluse}b, or weakly wrong like the Object Detector output in Fig.~\ref{fig:overview_tooluse}d.
In fact, the severity of a tool error depends on how critically the mistake impacts downstream task performance.  
In Fig.~\ref{fig:overview_tooluse}d, the Object Detector misidentifying the \texttt{Tomato} as an \texttt{Apple}, is crucial to the task, but mistaking objects like \texttt{Bread} would not hinder the task as much.
As the high-level goal is task success rather than perfect tool utilization, it is important to rectify critical mistakes, whereas harmless mistakes can be disregarded. 

To formalize this notion of ``task-critical'' tool-use mistakes,
we introduce an error threshold $\epsilon$ to define a range of tool outputs that are not ``critically'' wrong. Intervention is only necessary when the deviation between the tool output and the oracle, $d(o, o^*)$, is larger than $\epsilon$, thereby degrading the performance/quality of the final task output $\hat{y}$.

\vspace{-10pt}
\begin{align}
\label{eq1}
    d(o, o^*) &> \epsilon \; \Longrightarrow \; s_\text{task}(\hat{y} | o) < s_\text{task}( \hat{y}| o^*) \\
    & \text{where } s_\text{task} \coloneqq \text{task performance metric} \nonumber 
\end{align}
This is analogous to how humans approach errors; the goal is not a perfect world model but to accomplish a task.
As long as we can grab the apple, we do not need to know its exact shape or coordinates.

\section{Error sources}
\label{sec:error_sources}

The tool output $o$ is accurate if and only if:
\vspace{-1pt}
\begin{enumerate}
\setlength{\itemsep}{2pt}
\setlength{\parskip}{0pt}
    \item The tool inputs $i$ are accurate.
    \item The context $c$ is correct and sufficient.
    \item The tool $t_\theta$ makes correct predictions.
\end{enumerate}
Formally, to obtain $o$ with deviation smaller than $\epsilon$, $d(o, o^*)$, is a union of component error bounds:
{
\fontsize{10pt}{12pt}\selectfont{}
\begin{align}
\label{eq2}
    &d(o, o^*) < \epsilon\\
    &\Leftarrow \underbrace{d(i, i^*) < \epsilon_i}_{\text{tool input}} \; \land \;
    \underbrace{d(c, c^*) < \epsilon_c}_{\text{context}} \; \land \;
    \underbrace{d(t_\theta, t_{\theta^*}) < \epsilon_t}_{\text{tool correctness}} \nonumber
\end{align}
}
If any condition above is not met, output errors will lead to task failure. 
The following sections discuss each condition, and a table of corresponding real-world error scenarios is presented in App. \ref{sec:appendix_overview}.

\subsection{Input: $d(i, i^*) > \epsilon_i$} 
Imperfect tool inputs often result from incorrect outputs from a prior tool, like errors in LLM-generated code or noisy images. 
For deterministic tools (e.g., code interpreters), most errors are due to tool inputs, and 
malformed inputs typically trigger an error message.
However, well-formed inputs with incorrect content (e.g., ambiguous queries for search APIs) can produce erroneous outputs that inadvertently propagate through subsequent steps.

\subsection{Context: $d(c, c^*) > \epsilon_c$}
Partial observability of the surrounding environment can be another source of tool error, resulting in a lack of context for a tool to function properly. This is often inevitable early in the planning trajectory in interactive task settings. For example, an embodied agent may need to explore hidden objects in closed receptacles through trial-and-error, in order to obtain enough information for the task.

\subsection{Tool: $d(t_\theta, t_{\theta^*}) > \epsilon_t$} 

Tools themselves can make mistakes, even when the input or context is perfect. This situation is especially prominent as learnable tools are becoming more widely adopted in practice.
LLMs are prone to generating factually incorrect statements even when reference documents are provided through context~\citep{krishna2024genaudit}. Search APIs might fail not because of the input query's clarity, but due to an imperfect database/dense retrieval method. 
The tool's precision can also contribute to failure -- heuristic-based search/manipulation robot policies can fall apart when they lack the precision needed to address the complexity of real-world scenarios.

Due to the absence of explicit error signals, tool-based errors require the tool-using model to reason over indirect cues. In easier cases, errors can be recognized based on well-calibrated confidence scores.
Much harder cases, however, arise when a tool confidently produces errors.
In such scenarios, a broader context may help identify these hidden errors. Multiple tools presenting conflicting evidence (e.g., fact verification tool vs search API), disagreement between different modalities~\citep{crossmodal-compensation}, or prediction inconsistencies over multiple trials~\citep{Kadavath2022LanguageM,LLM-SC-Wang2023} or timesteps~\citep{CuriosityChaplot2020}, may help surface potential limitations of the tool. 
\looseness=-1
\section{Recovery behaviors}

Next, we categorize current recovery methods from previous literature into two groups: \textbf{Refine} and \textbf{Replace}, and advocate for meta-cognitive reasoning.

\subsection{Refine: $i \to i^*$, $c \to c^*$}
\label{sec:refine}

Recovering from tool failures often involves refining the tool input. This is particularly effective when the failure is followed by explicit feedback signals that indicate ``what'' to fix. Inputs can be rewritten guided by
API error messages and human/LLM feedback~\citep{2023madaan-self-refine,Shinn2023ReflexionLA,wang2023mint}. In the planning literature (e.g., TAMP~\cite{garrett2021integrated, ding2023task}), this is referred to as ``closed-loop planning,'' where plans are continuously updated by new observations, task progress, or clarification questions~\citep{inner-monologue2022Huang,progprompt2022Singh,Song2022LLMPlannerFG}. Augmenting the context based on increased observability changes the input's interpretation. 
Refine methods are well-suited to LLMs as they can flexibly accept varying lengths of text-based feedback. In contrast, corrections to other modalities (e.g. image lighting or non-verbal communication) remain open challenges for VLMs.

\looseness=-1
\subsection{Replace: $t_\theta \to t_{\theta^*}$}
\label{sec:replace}
When errors originate from the tool itself, the aim is to move $t_\theta$ closer to $t_{\theta^*}$, aligning it more closely with the final task. Mitigation strategies vary based on how easily the tool can be fixed at inference time. For LLMs, in-context examples are used to elicit specific task capabilities from more generic reasoning abilities, a method further enhanced by retrieving samples that are more pertinent to the specific test example~\citep{rubin-etal-2022-learning,Song2022LLMPlannerFG}. 
Ensembles over multiple predictions also offer a non-invasive way to improve tool performance
~\citep{Anil2023GeminiAF,LLM-SC-Wang2023,
chen2024more-llm-calls}.
Test-time adaptation methods~\citep{wang2021tent} can be useful, though application requires access to the tool's internal parameters. 
The aforementioned strategies focus on improving the tool's performance in isolation, which may not translate to better task performance. In Fig.~\ref{fig:overview_tooluse}d, better ImageNet performance does not guarantee detecting the Tomato. 
Understanding the interplay between tools and task performance remains an open question of system dynamics and credit assignment.

When improving the tool is not viable or when adjustments are insufficient, the best strategy can be to switch to a different tool. Research on assistance-seeking agents implicitly model this behavior, with agents identifying when
to delegate the action to a human/oracle~\citep{singh2022ask4help,xie2022ask}. In NLP, \citet{krishna2024genaudit} introduce a fact-checking tool that edits unsupported claims in LLM-generated summaries, advocating for the strategic use of alternative tools to ensure quality and reliability.

\subsection{LLMs as a Meta-Reasoner: $\epsilon_i, \epsilon_c, \epsilon_t \uparrow$}

For humans, the tools we employ are not perfect. But tools can err because humans can fix incorrect outputs -- misrecognized card numbers through an OCR system are corrected ad-hoc by the user. Similarly, imbuing LLMs with the ability to recognize and handle errors flexibly allows for tools to make mistakes, effectively increasing the permissible error thresholds of the tool components $\epsilon_i, \epsilon_c, \epsilon_t$ in Eq. \ref{eq2}. 
An LLM's meta-cognitive ability to reason over uncertainty and realize its knowledge limits have received some attention~\citep{Kadavath2022LanguageM,kuhn2023semantic}. The next step is to jointly reason over their uncertainty/knowledge and that of another tool or agent. This compounds in multi-tool or multi-LM settings.
Existing recovery methods that presuppose the cause and tweak a single knob may not yield overall improvement unless limitations of the right variables are resolved. 

\vspace{0.25em}
\noindent In summary, we identify three challenges:
\vspace{-0.5em}
\begin{enumerate}[noitemsep]
    \item \textbf{Failure Detection}: Recognizing failures and assessing their severity -- $d (o, o^*) > \epsilon$ ? 
    \item \textbf{Fault Assignment}: Identifying which tool caused the error (in multi-tool settings), with the exact source -- $i$, $c$, or $t_\theta$?
    \item \textbf{Recovery Planning}: Selecting the most effective recovery strategy. 
    Refine vs Replace?
\end{enumerate}
\vspace{-0.5em}
Explicit error signals (though rare) can obviate all three problems. 
More importantly, silent tool errors are the opposite case, where even detection is not straightforward although the problem is pervasive. In this work, we delve into ``silent'' tool errors, a relatively overlooked area in tool-error research, focusing on the foremost problem: error detection. 

\looseness=-1
\section{A broken calculator}
\label{sec:broken_calc}
Humans use tools with a rough expectation of what correct results should look like, allowing them to spot outputs that are obviously wrong. For example, for multiplying 120 by 131, we can expect a result around 10,000 and ending in zero, even if we don't know the exact answer.
If the tool makes arithmetic mistakes, 
can LLMs also detect faulty outputs?

\subsection{Task setting}
\label{sec:calc_task_setting}
We devise a controlled setting where an LLM answers simple math problems with an external tool, a calculator. In this case, the calculator is broken and returns incorrect outputs. 

First, we programmatically generate 300 equations that involve two random operators from $\{+, -, \times\}$ and three random integers (e.g., $9 \times (20 + 7)$). The equations have three levels of difficulty, which is determined by the range that the integers are sampled from: easy $[-20, 20]$, medium $[-100, 100]$, and hard $[-1000, 1000]$. 
We give the incorrect tool output to the model, and test whether models are able to recognize the error. We compare five different models: GPT-3.5 and GPT-4, Command-R and Command-R+, Gemini-1.5.

\begin{figure}[t]
\centering
\begin{lstlisting}
# Task
What is the answer to: (2 + 3) * 5?

(*@\textbf{Refer to the tool output below.}@*)
(*@\textbf{\# Calculator API}@*)
(*@\textbf{result = (2 + 3) * 5}@*)
(*@\textbf{result}@*)
(*@\textbf{25}@*)  # broken tool setting -> (*@\textcolor{c4}{21}@*) / (*@\textcolor{c2}{205}@*) / (*@\textcolor{c1}{-25}@*)

# Format
Return your answer in this format:
Thought: Your reasoning process
Answer:
...

# Answer
\end{lstlisting}
\vspace{-10pt} 
\caption{Prompt for a math problem using tool outputs. The result \texttt{25} is perturbed in the Broken scenario: \textcolor{c4}{Digit replacement}, \textcolor{c2}{Magnitude shift}, or \textcolor{c1}{Sign inversion}.}
\label{prompt:math_1}
\end{figure}

\subsection{Preliminary experiments}
\label{sec:prelim_exp}
We begin by estimating the models' capabilities to solve math problems on their own, to better understand the downstream implication of having a credible/broken calculator in the loop. Specifically, we query the LLM with five different prompts -- three non-tool and two tool-use prompts.

\paragraph{Non-tool setting} The non-tool settings serve as a proxy to gauge the model's task capability, providing a basis to compare the effects of incorporating tools with varying levels of credibility. We ask the model to solve the math problems on its own, with three different prompting methods:
\vspace{-0.5em}
\begin{enumerate}[noitemsep,leftmargin=15pt]
    \item Direct: Asking the equation directly (e.g., \texttt{"What is the answer to (2+3)*5?"})
    \item Chain-of-Thought (CoT): Asking to explain its reasoning step-by-step prior to answering.
    \item CoT Few-Shot: In addition to reasoning, the model is provided five in-context examples.
\end{enumerate}

\paragraph{Tool-use setting} 
\label{sec:prelim-tool-use}
We assume two types of calculators -- Correct and Broken. Fig.~\ref{prompt:math_1} shows the tool-use prompt, where the model is asked to answer the question referring to the tool output (\textbf{bold}). For Correct tool, the ground truth answer is provided as the tool result.
For Broken tool, we give a perturbed answer using one of the following three:
\vspace{-0.5em}
\begin{enumerate}[noitemsep,leftmargin=15pt]
    \item Digit replacement: One digit is replaced with a different number (e.g., $25 \to 21$)
    \item Magnitude shift: Digits are inserted/removed, resulting in magnitude shifts in the range $10^{-2}$ and $10^3$ (e.g., $25 \to 205$)
    \item Sign inversion: The sign is flipped, changing positive numbers to negative and negative numbers to positive (e.g., $25 \to -25$)
\end{enumerate}
Inspired by~\citet{wei2022chain,yao2023react}, we specify a ``Thought'' section, to encourage the model to generate its reasoning prior to answering.

\paragraph{Results}

\begin{figure}[!t]
    \centering
    \includegraphics[width=1\linewidth,trim={0 0 0 0},clip]{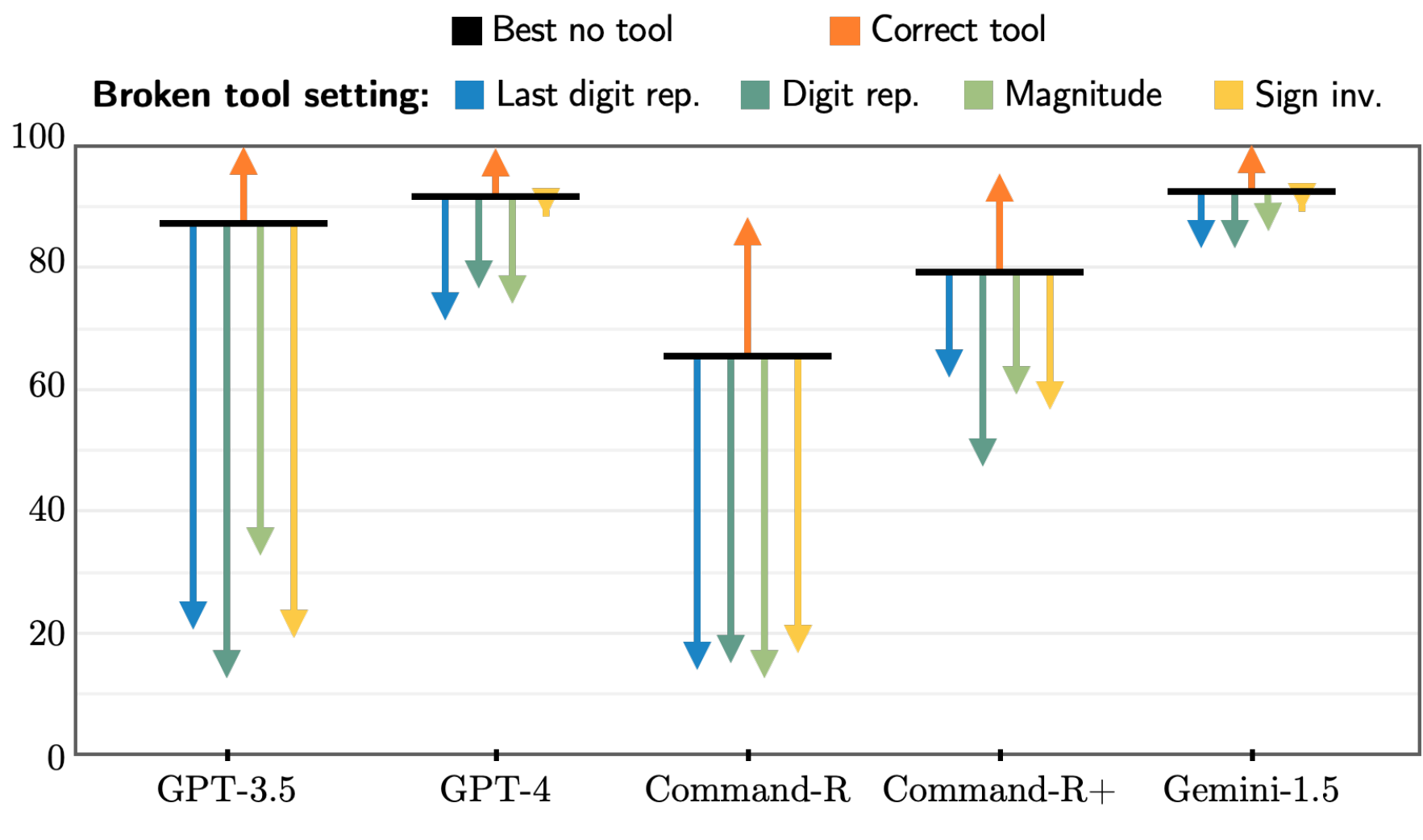}
    \vspace{-2em}
    \caption{Math accuracy of models. The black bar indicates the best accuracy \textit{without} tool-use; upward orange/downward arrows respectively indicate performance with correct/broken tool-use.}
    \label{fig:broken_tool}
\end{figure}

We report the preliminary experiment results in App. \ref{app:math_problem} and Fig.~\ref{fig:broken_tool}. 
When the tool is broken, the accuracy drops drastically for all perturbation categories, with the exception of Sign Inversion on GPT-4 and Gemini-1.5.
With broken tools, performance drops far below the best no-tool setting's performance, up to 47\%. We find that models tend to overtrust tools -- copying the incorrect output (with hallucinated justification) rather than ignore the tool in favor of its own answer.

\subsection{In-context intervention strategies}
Humans leverage various contextual cues like prior tool failures to calibrate the level of trust associated with their tools. Further, AI chatbots include disclaimers like ``The model can make mistakes'' to ensure answers are scrutinized.
Can LLMs also leverage such information effectively? 

We test three types of contextual cues that can raise the awareness towards potential tool mistakes: a simple disclaimer, prediction confidence scores, and a checklist of criteria to look out for. For each method, we evaluate the prediction accuracy on both perturbed and non-perturbed tool outputs, in ZST, CoT, and FST settings. The prompt...

\noindent\textbf{Oblivious (Obl.)} does not mention any indications that the tool can cause errors Fig.~\ref{prompt:math_1}.

\noindent\textbf{Disclaimer (Disc.)} includes a simple disclaimer: \textit{``The tool can sometimes give incorrect answers. Please verify the correctness of the tool output.''}

\noindent \textbf{Confidence (Conf.)} includes the confidence score of the tool's prediction, in addition to the disclaimer. Since the calculator is not a probabilistic model, we devise a score [0,1] based on the string edit distance between the ground truth and the perturbed output. 
For learned tools, model confidence can be used. 

\noindent \textbf{Checklist (Check.)} is motivated by heuristics that humans use, which includes a list of criteria to check the tool output, based on the perturbation. For the math task, the checklist consists of:
\vspace{-0.5em}
\begin{enumerate}[noitemsep]
    \item Is the positive or negative sign correct?
    \item Is the magnitude of the number correct?
    \item Is the last digit correct?
    \item Are all the digits correct?
\end{enumerate}
\vspace{-10pt}

\begin{table}[t]
    \centering
    \small
 \begin{tabular}{@{}l@{\hspace{5pt}}c@{\hspace{4pt}}c@{\hspace{4pt}}c@{\hspace{4pt}}c@{\hspace{1em}}c@{\hspace{4pt}}c@{\hspace{4pt}}c@{\hspace{4pt}}c@{\hspace{1em}}c@{\hspace{4pt}}c@{\hspace{4pt}}c@{\hspace{4pt}}c@{}}
    \toprule
    &&\multicolumn{3}{c}{\textbf{ZST}}&\multicolumn{4}{c}{\textbf{CoT}}&\multicolumn{4}{c}{\textbf{CoT+FST}} \\
    Model 
    & \rotatebox{90}{Obl.} & \rotatebox{90}{Disc.} & \rotatebox{90}{Conf.} & \rotatebox{90}{Check.} 
    & \rotatebox{90}{Obl.} & \rotatebox{90}{Disc.} & \rotatebox{90}{Conf.} & \rotatebox{90}{Check.} 
    & \rotatebox{90}{Obl.} & \rotatebox{90}{Disc.} & \rotatebox{90}{Conf.} & \rotatebox{90}{Check.} \\
    \midrule
    GPT-3.5 & 23 & \textbf{53} & 44 & 46 
            & 46 & \textbf{81} & 79 & 80 
            & 87 & \underline{\textbf{89}} & 86 & 84 \\
    GPT-4   & 76 & 82 & \textbf{85} & \textbf{85} 
            & 86 & 89 & 89 & \underline{\textbf{91}} 
            & 90 & \textbf{91} & 88 & 89 \\
    Command-R & 16 & 14 & \textbf{16} & 14 
            & 29 & 42 & 44 & {\textbf{47}}
            & 11 & 23 & \underline{\textbf{53}} & {46} \\
    Command-R+ & 57 & 76 & {79} & \textbf{81}
            & 60 & \textbf{84} & 82 & 76
            & 71 & 82 & \underline{\textbf{86}} & 78 \\
    Gemini-1.5 & 84 & \textbf{90} & 76 & 87
            & 93 & \textbf{95} & 95 & 90 
            & \textbf{94} & \textbf{94} & 94 & 94 \\
    \bottomrule
\end{tabular}
\vspace{-5pt}
\caption{Accuracy of models on math equations with in-context intervention methods against broken tools}
\label{tab:math_acc2}
\end{table}
\paragraph{Results}
\label{sec:broken_calc_results_1}
Table \ref{tab:math_acc2} shows how effectively each method helps the LLM notice and correct mistakes. For most models, even a simple disclaimer prevents naively believing perturbed answers, boosting accuracy up to 30\%. As humans, LLMs can better detect mistakes when provided the context that tools can be wrong. Chain-of-thought prompting and in-context examples further help models recover performance, nearly to the best no-tool scores.

\section{Detecting tool-based mistakes}
The results in \S\ref{sec:broken_calc} suggest that it is challenging for LLMs to simultaneously detect and override faulty outputs, even for capabilities that are decently performed without tools. Thus, next we narrow the LLM's responsibility to ``detecting'' mistakes.\footnote{
We reformulate the calculator setting into a binary Accept/Reject task (Fig.~\ref{prompt:math_2}). 
We balance the 300 perturbed equations in \S\ref{sec:prelim-tool-use} with 300 correct samples to account for false positives.}

\paragraph{Results}
\begin{table}[!tbp]
    \centering
 \small
 \begin{tabular}{l@{\hspace{2em}}c@{\hspace{6pt}}
    c@{\hspace{6pt}}c@{\hspace{6pt}}c@{\hspace{2em}}
    c@{\hspace{6pt}}c@{\hspace{6pt}}c@{\hspace{6pt}}c}
    \toprule
    &
    \multicolumn{4}{c}{\textbf{ZST}}& 
    \multicolumn{4}{@{}c}{\textbf{CoT}}\\
    Model &
    \rotatebox{90}{Obl.} & \rotatebox{90}{Disc.} & \rotatebox{90}{Conf.} & \rotatebox{90}{Check.} & 
    \rotatebox{90}{Obl.} & \rotatebox{90}{Disc.} & \rotatebox{90}{Conf.} & \rotatebox{90}{Check.} \\
    \midrule
    GPT-3.5 & 79 & \underline{\textbf{86}} & \underline{\textbf{86}} & 83 & 70 & 67 & \textbf{83} & 75 \\
    GPT-4 & 92 & \textbf{95} & 94 & 91 & 96 & \underline{\textbf{97}} & 96 & 94 \\
    Command-R & 62 & 64 & \textbf{67} & 60 & 59 & 68 & \underline{\textbf{80}} & 71 \\
    Command-R+ & 83 & \underline{\textbf{89}} & 87 & 77 & 73 & 78 & \textbf{81} & 77 \\
    Gemini-1.5 & 92 & 94 & 94 & \underline{\textbf{96}} & 95 & \underline{\textbf{96}} & \underline{\textbf{96}} & 89 \\
    \bottomrule
\end{tabular}
\vspace{-5pt}
\caption{Accuracy of models on the Accept/Reject task on calculator outputs.
}
\label{tab:math_binary}
\end{table}
The models are often able to identify the incorrect outputs (Table \ref{tab:math_binary}) despite not being able to produce the correct answer -- even in conditions where they would have without a tool present. Smaller models (GPT-3.5, Command-R) are more sensitive to in-context information.
Where in Oblivious, most small model errors are due to overtrusting tools, and with in-context intervention, the prediction skews heavily towards rejecting outputs, leading to high false positive rates. In contrast, errors occur in similar rates for the larger models.

\begin{figure*}[!t]
    \centering
    \includegraphics[width=\linewidth]{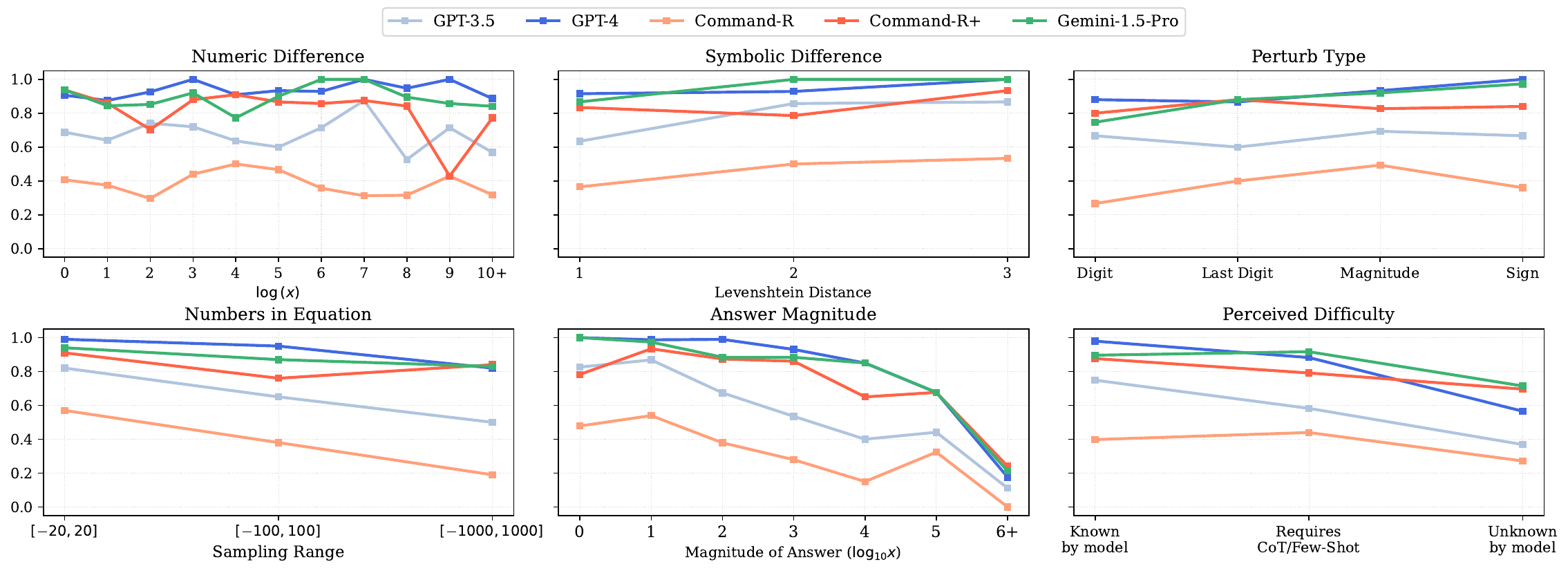}
    \vspace{-20pt}
    \caption{The rejection rate on the perturbed calculator outputs with respect to six features.}
    \label{fig:rejection_feature_correlation}
\end{figure*}

Surprisingly, CoT does not always improve performance over Zero-shot. We find that the majority of CoT errors are the model falsely rejecting correct outputs -- caused by failure in faithfully copying the original equation's terms in its reasoning steps.
Incorrect reasoning cases are more frequently observed in the CoT setting, contradicting Table \ref{tab:math_acc2} where CoT outperformed Zero-shot. While more investigation is needed, we speculate that the effectiveness of CoT might depend on task complexity, because the model is burdened to simultaneosly 1. solve the equation and 2. spot mistakes in the Detection+CoT setting. A two-step process where the LLM first generates its answer, then compares its own answer to tool outputs in a second call may alleviate this issue, which we leave to future work.

\subsection{When are mistakes easier to detect?}
For humans, whether a mistake is detected might depend on the type of mistake (blatant vs subtle), the difficulty of the original question, or the answerer's task proficiency. Are some mistakes, past a certain level of deviation, just more obvious than others? Does the property of the question matter? Or does it relate to the model's internal knowledge -- do you need to ``know'' the answer to detect errors? 
In Fig.~\ref{fig:rejection_feature_correlation}, 
we analyze the models' rejection rate on the perturbed outputs with respect to six features:
\begin{itemize}[itemsep=1pt,leftmargin=0pt]
\item[] \textbf{{Numeric Difference}} The absolute difference between the correct and perturbed answer. 
\item[] \textbf{{Symbolic Difference}} The string edit (Levenshtein) distance. Smaller symbolic deviations are expected to be less noticeable. Symbolic difference only loosely correlates with numeric differences ($\rho=0.49$), for example $123$ to $-123$ vs $119$.

\item[]\textbf{Perturbation Type} Digit replacement, Magnitude shift, and Sign inversion from \S\ref{sec:prelim-tool-use}. We separate last digit replacement as it is easier for humans to detect than other digit positions by mental math.

\item[]\textbf{Magnitude in Equation} Equations are binned into three difficulty levels (\S\ref{sec:calc_task_setting}), based on the magnitude of numbers involved in the equation. Relatedly, LLMs have been shown to find larger numbers harder to reason over~\cite{nogueira2021investigatinglimitations,lee2023teaching, an2023placevalue,duan2024from}.

\item[]\textbf{Answer Magnitude} The magnitude of the correct answer, in log scale $(\log_{10}{|x|})$. Similar to above, but provides more fine-grained measurements.

\item[]\textbf{{Perceived Difficulty}} This is inferred via the model's ability to answer the equation in \S\ref{sec:prelim_exp}. The categories are: The model (1) answered correctly with a ``Direct'' prompt, (2) required CoT or Few-Shot examples, and (3) gets the equation wrong even after applying these methods. The number of samples vary for each bin, depending on the model.
\end{itemize}
\vspace{-2pt}
\looseness=-1
Numeric/String Difference and Perturbation Type attribute the rejection rate to the error's ``wrongness.'' Magnitude is associated with the question itself, and Perceived Difficulty targets the model's internal knowledge.

\subsection{Analysis}
\begin{itemize}[itemsep=1pt,leftmargin=0pt]
\item[]\textbf{Numeric vs Symbolic} Unlike 
numeric difference, 
symbolic deviations appear highly correlated with rejection rates. This aligns with literature that LLMs are not performing arithmetic ``reasoning,'' but memorizing strings~\cite{chang2024language}.
\item[]\textbf{Perturbation Types} For humans, Sign Inversion and Last Digit are likely the easiest to spot. 
LLMs also find some perturbation types more obvious than others -- 
Sign Inversion for GPT-4 and Gemini, Magnitude for Command-R and GPT-3.5, and Last Digit Replacement for Command-R+. Most models find Last Digit Replacements easier to spot than other digits. Sensitivity is likely attributable to differing representations/tokenization
~\cite{nogueira2021investigatinglimitations, liu2023goatfinetunedllama}.
\item[]\textbf{Large Numbers} Models struggle with large values in both Numbers in Equation and Magnitude. 
Equations with large numbers can be easier depending on the operations involved. For instance, $(1000-998)\times 2 = 4$ is easier than $10\times 11\times 12 = 1320$. Notably, the rejection rate for answers larger than $10^6$ drops sharply for all models.
\item[]\textbf{Perceived Difficulty} Problems that are more easily answered by the model, are also more easily detected when exposed to errors.
While this might raise a question on the utility of imperfect tools, we find that the larger models (GPT-4, Gemini-1.5-Pro, Command-R+) can ``detect'' the mistake for the majority of questions, even for ones that it were not able to answer correctly. This sheds light on the feasibility of using LLMs as a tool planner that evaluates the credibility of tools and reroutes functions accordingly to alternative tools. Smaller models, however, overtrust the tool and allow errors to pass.
\end{itemize}

\section{Natural tool errors: ALFRED}
\begin{figure}[!tbp]
    \centering
    \includegraphics[width=1\linewidth]{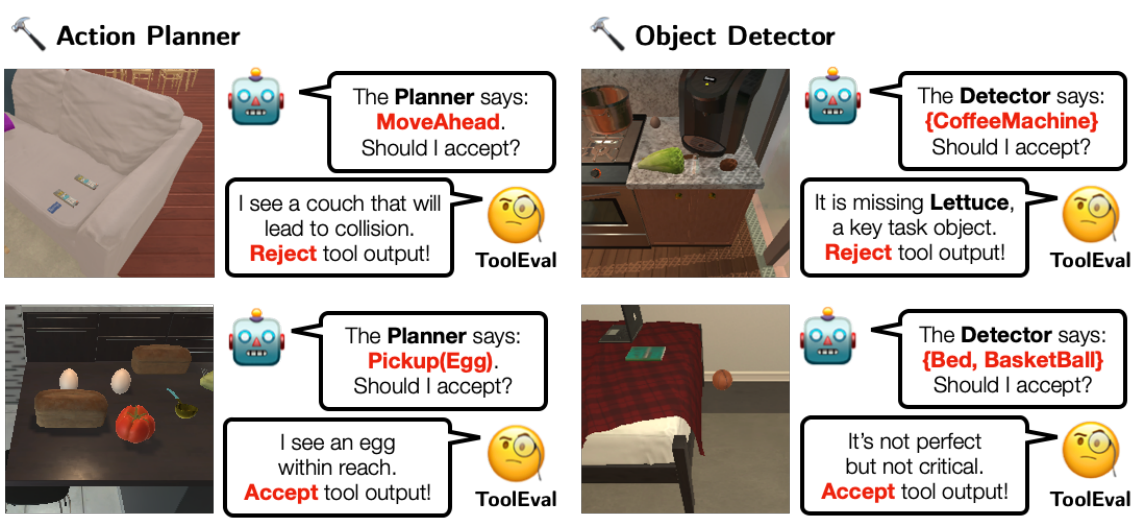}
    \vspace{-2em}
    \caption{Evaluating two tool outputs in ALFRED -- Action Planner (Left) and Object Detector (Right). The LLM is asked whether to Accept/Reject the tool output, based on the provided image and task context.}
    \label{fig:alfred_tool_examples}
\end{figure}

We now consider a setting where tool-based errors occur more naturally via ALFRED~\cite{ALFRED20}, an embodied instruction following benchmark. 
Involving language understanding, perception, spatial reasoning, and action planning capabilities, a common approach is to incorporate multiple specialized modules~\cite{pmlr-v164-blukis22a,FILM}, as opposed to end-to-end training. 

Multiple modules, or tools collaborating with each other in ALFRED offer a unique opportunity to study the robustness of LLMs to various tool errors. As in Fig.~\ref{fig:overview_tooluse}d, the object detector's mistakes are silently passed on to subsequent tools, leading to error cascades in the Action Planner. In such scenarios, LLMs that can detect tool errors help improve the system's robustness, by correcting some obvious semantic anomalies~\cite{Elhafsi2023-kq} or delegating operations to other tools or humans.

In this section, we investigate whether LLMs can detect these realistic, multimodal tool errors arising from individual modules used in the FILM architecture~\cite{FILM}. Specifically, we test the LLM's fault detection capability on two distinct tools -- the object detector and the action planner.\footnote{Object detection uses a finetuned MaskRCNN model. Action planning is done by the Fast Marching Method~\cite{fmm-planner}, a heuristic-based algorithm.}

\subsection{Multimodal tool-error detection dataset}
We create a classification task where the model Accept/Rejects the tool output, based on the current context. 
For the action planner, the model has to assess the feasibility of the predicted action, and reject actions that are to fail (e.g., facing an obstacle for \texttt{MoveAhead}, Fig.~\ref{fig:alfred_tool_examples}).
For the object detector, the LLM evaluates the correctness of the detection results with respect to the image, and reject outputs that mistaken important task objects. We note that imperfect outputs can still be labeled as ``Accept'' if only containing task-irrelevant errors.

We collect agent trajectories from the ALFRED validation set with actions and API responses whether the action succeeded.
For the object detector, we gather RGB images with detection results and groundtruth semantic information. We provide detailed statistics of each dataset in App. \ref{app:alfred_dataset}.

\begin{table}[t]
    \centering
 \small
 \begin{tabular}{p{1.5cm}@{\hspace{5pt}}c@{\hspace{1em}}c@{\hspace{5pt}}
    c@{\hspace{5pt}}c@{\hspace{5pt}}c@{\hspace{2em}}
    c@{\hspace{5pt}}c@{\hspace{5pt}}c@{\hspace{5pt}}c}
    \toprule
    &
    \textbf{VLM}&
    \multicolumn{4}{c}{\textbf{ZST}}& 
    \multicolumn{4}{@{}c}{\textbf{CoT}}\\
    & & 
    \rotatebox{90}{Obl.} & \rotatebox{90}{Disc.} & \rotatebox{90}{Conf.} & \rotatebox{90}{Check.} & 
    \rotatebox{90}{Obl.} & \rotatebox{90}{Disc.} & \rotatebox{90}{Conf.} & \rotatebox{90}{Check.} \\
    \midrule
\multirow{2}{\linewidth}{\textbf{Action Planner}} &
GPT-4o & 43 & 42 & 40 & \textbf{44} & 57 & 55 & 52 & \underline{\textbf{60}} \\
& Gemini & 49 & 55 & 50 & \textbf{63} & 64 & 64 & 62 & \underline{\textbf{65}} \\
\midrule
\multirow{2}{\linewidth}{\textbf{Object Detector}} &
GPT-4o & \textbf{68} & {\textbf{68}} & 66 & 67 & 68 & \underline{\textbf{69}} & 66 & 68\\
& Gemini & 60 & 60 & 56 & \textbf{62} & \underline{\textbf{67}} & 66 & 65 & 66 \\
    \bottomrule
\end{tabular}
\vspace{-5pt}
\caption{F1 score on the Accept/Reject task on two tool outputs in ALFRED. We compare interventions (Disclaimer, Confidence, Checklist) with ``Oblivious.''}
\label{tab:alfred_result}
\end{table}

\subsection{Experimental setting}
\paragraph{Models} We test tool evaluation accuracy against the two best closed-source Vision-Language Models: GPT-4o and Gemini-1.5-Pro-latest.
As in the calculator, we evaluate models on Zero-Shot (ZST) and Chain-of-Thought (CoT) settings. The prompt includes the task state (e.g., current subgoal, steps taken), tool docstrings (e.g., possible actions, object categories), and the current tool output. We provide example prompts in the Appendix: Action Planner (\ref{app:alfred_action}), Object Detector (\ref{app:alfred_perception}).

\subsection{Results}
Models are able to reach 60-70 F1 scores with raised awareness through ICL and CoT prompting (Tab.~\ref{tab:alfred_result}).
In particular, specifying the potential failure modes in the Checklist prompt is effective for evaluating the action planner, where the error modes are more diverse than the Object Detector. 
In contrast, giving the raw confidence scores is not as helpful, 
as it demands additional interpretation.
As these results are all zero-shot evaluations, we expect further improvements in few-shot or finetuning scenarios.
Details of the Action Planner and Object Detector along with analysis are presented in Appendix \ref{app:alfred}.

\section{Conclusion}
We characterize the trust dynamics of modern LLMs with respect to tool usage. By establishing an extensive taxonomy of tool-related errors and recovery strategies, we identify fundamental challenges associated with integrating learned tools. Our experiments span both synthetic and natural tool failures, and affirms current LLMs' ability to identify silent tool failures. This work paves the way for future research on harnessing LLMs as sophisticated tool-reasoners.

\newpage
\section{Limitations}
This study, while comprehensive in its scope, has certain limitations regarding the diversity and breadth of the models and datasets used. Firstly, for the calculator experiments, we employ five LLMs, mostly closed-source. Including smaller, open-source models, and models specifically fine-tuned for tool-use would have offered more insights into the models' tool trusting behavior. In the experiments involving embodied agents, we limited our focus to only two API-based Vision-Language Models (VLMs). Incorporating smaller, open-source VLMs would have offered opportunities to explore into the models' internal workings, revealing additional nuances in how models handle unreliable tools.

Secondly, the action planner and object detection dataset we constructed based on ALFRED trajectories is fairly small in size -- Action Planner (490) and Object Detector (214). In terms of diversity, running multiple models/agents in addition to FILM would have enabled collecting a wider array of failure modes. Moreover, the action's success or failure is highly dependent on the affordances provided by the AI2-THOR framework which may not accurately reflect real-world scenarios. For example, a `Put' action might fail due to the system perceiving a surface as cluttered, even when there is visibly sufficient space available. A dataset encompassing a wider variety of scenarios and higher diversity would potentially provide deeper insights into the practical applications and limitations of current AI systems in navigating real-world environments.

\newpage
\bibliography{acl}

\clearpage
\newpage
\appendix

\renewcommand \thepart{}
\renewcommand \partname{}
\addcontentsline{toc}{section}{Appendix} %
\part{Appendix} 
\parttoc 
\label{sec:appendix}

\section{Overview of Tool Errors}
\label{sec:appendix_overview}

In \autoref{tab:tool_errors}, we compile a list of tools that support various modalities, with respective real-world tool-error scenarios. We categorize specific error scenarios by its source of failure -- the tool input, the tool itself, or context information. 

\begingroup
\begin{table*}[t]
\centering
\scriptsize
\renewcommand{\arraystretch}{1.5} 
\begin{tabular*}{\textwidth}{l|P{0.7in}|p{0.8in}|P{1.0in}P{1.0in}P{1.0in}}
    \toprule
        \textbf{} &\textbf{} &\textbf{} &\multicolumn{3}{c}{\textbf{Source of failure}} \\
        \textbf{Modality} &\textbf{Capability} &\textbf{Tool} &\multicolumn{1}{c}{\textbf{Tool input}} &\multicolumn{1}{c}{\textbf{Tool itself}} &\multicolumn{1}{c}{\textbf{Context}} \\
    \midrule
        Text & Mathematical \newline computation &Calculator \newline Code interpreter &\tabitem API syntax error \newline \tabitem Incorrect content & NA & NA \\ 
            &Code \newline validation&Code interpreter & \tabitem Code syntax error \newline \tabitem Version updates (e.g., deprecated functions) \newline \tabitem Incorrect content & NA & NA \\
            &World \newline knowledge &Search API & \tabitem Ambiguous query & \tabitem Incomplete DB  \newline \tabitem Irrelevant results (e.g., different word sense) & \\
            &Task planning &LLM/VLM & \tabitem Prompt includes non-existent objects due to previous perception errors & \tabitem API call failure \newline \tabitem Plan includes unsupported actions/objects \newline \tabitem Incorrect steps & \tabitem Invalid plan due to partial observability (e.g., closed receptacles) \\
    \midrule
        Image &Text recognition & OCR model & \tabitem Blurry/noisy image & \tabitem Parsing mistakes & \\
            &{Visual \newline perception} &Vision-Language Models (CLIP) \newline Semantic segmentation (Fast-RCNN) \newline Object detectors (M-DETR) & \tabitem Camera noise  \newline \tabitem Poor lighting \newline & \tabitem Unknown object \newline \tabitem Detection failure \newline \tabitem Hallucination \newline \tabitem Wrong categories \newline \tabitem Bad segmentation mask & \\
            & & Depth estimators & & \tabitem Estimation errors & \\
    \midrule
        Sensory Perception  & Pose Estimation, Map building  &SLAM & \tabitem Sensor drift & \tabitem Algorithmic errors & \tabitem Environmental interference (e.g. moving humans, key object change) \\
    \midrule
        Audio &{Auditory \newline perception} &Speech-to-text API (Socratic Models) & \tabitem Audio noise & \tabitem Recognition errors & \\
    \midrule
        Action &Navigation & Path-planning algorithms (A*, Fast Marching Method)  & & \tabitem Collision \newline \tabitem Circling with no progress & \tabitem Change in obstacle locations \newline 
         \\
            & Manipulation & Skills &  & \tabitem Grip failure \\
    \bottomrule
\end{tabular*}
\caption{\textbf{Overview of Tool Errors.} API syntax errors are a shared case of input-based failures across tools. Similarly, network issues are shared across tools as environmental failures.}
\label{tab:tool_errors}
\end{table*}
\endgroup

\section{Math problems}
\label{app:math_problem}

\autoref{tab:math_acc1} reports the accuracy of models on ``answering'' math equations, plotted in \autoref{fig:broken_tool}. The numbers in the parentheses indicate the relative gain/loss compared to the best no-tool setting (in \textbf{bold}). In short, Chain-of-Thought prompting improves arithmetic performance, which is further enhanced by few-shot in-context examples. 
Correct tool-use yields strongest results, supporting existing literature that employ reliable tools. 

We share an example prompt for Accept/Reject task for the calculator setting in \autoref{prompt:math_2}. This is comparable to \autoref{prompt:math_1}, where the task inputs are identical, but the primary task is to ``answer'' the equation rather than ``evaluating'' the tool output.

\begin{table}[tbp]
    \centering
 \resizebox{0.48\textwidth}{!}{\begin{tabular}{l|ccc|c|c}
    \toprule
    Model & {Direct} & {CoT} & {CoT-FS} & {Correct tool} & {Broken tool} \\
\midrule
GPT-3.5     & 61.0 & 79.7 & \textbf{85.3} & 98.7 (+13.4) & 22.7 (-62.6)\\
GPT-4       & 64.0 & 89.0 & \textbf{89.7} & 97.7 (+8.0)& 76.0 (-13.7)\\
Command-R   & 34.3 & 52.3 & \textbf{63.3} & 86.3 (+23.0) & 16.0 (-47.3) \\
Command-R+  & 62.0 & 75.7 & \textbf{77.3} & 93.7 (+16.4) & 56.7 (-20.6) \\
Gemini-1.5  & 86.7 & \textbf{90.3} & 88.7 &  98.3 (+8.0) & 83.7 (-6.6) \\
    \bottomrule
\end{tabular}}
\caption{Average accuracy of models on math equations based on various prompting methods.}
\label{tab:math_acc1}
\end{table}

\begin{figure}[p]
\centering
\begin{lstlisting}
# Task
You are given the equation: (2 + 3) * 5. (*@{\textbf{The task is to}@*) 
(*@{\textbf{evaluate the result of the equation provided by the tool.}}@*)

Refer to the tool output below.
# Calculator API
result = (2 + 3) * 5
result
-25 # broken tool setting -> (*@\textcolor{c4}{21}@*) / (*@\textcolor{c2}{205}@*) / (*@\textcolor{c1}{-25}@*)

# Format
Return your answer in this format:
Thought: Your reasoning process
(*@{\textbf{Evaluation: Accept/Reject}@*)
...

# Answer
\end{lstlisting}
\caption{Example Accept/Reject prompt for the output of the calculator. The modified Fig.~\ref{prompt:math_1} instructions are in \textbf{bold}. We color-code the three perturbation methods as: \textcolor{c4}{Digit replacement}, \textcolor{c2}{Magnitude shift}, \textcolor{c1}{Sign inversion}.}
\label{prompt:math_2}
\end{figure}

\section{ALFRED}
\label{app:alfred}

\subsection{Dataset}
\label{app:alfred_dataset}

\begin{figure*}[h!]
    \centering
    \includegraphics[width=1\linewidth]{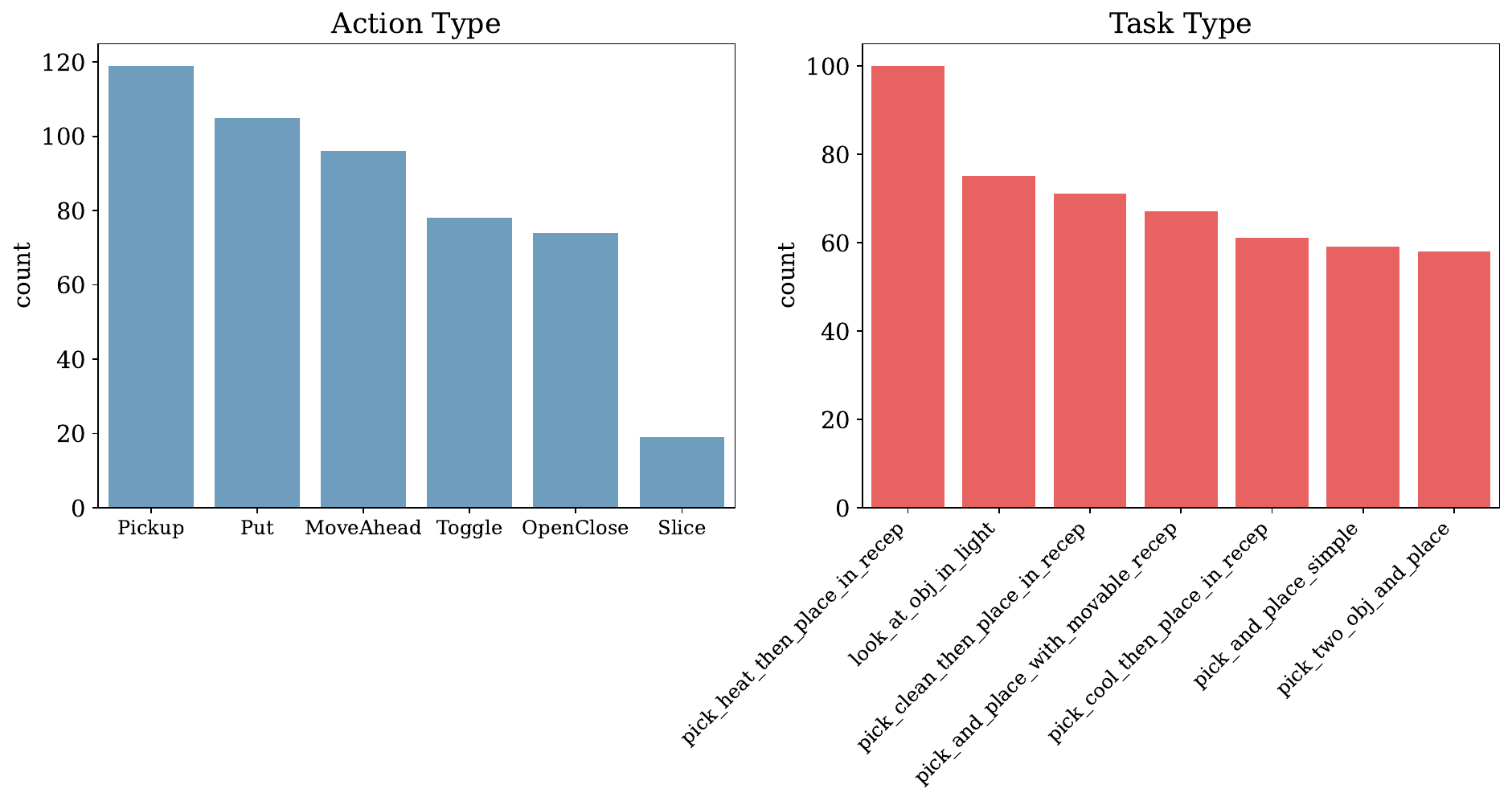}
    \caption{Histogram of actions (left) and task types (right) in the dataset}
    \label{fig:action_count}
\end{figure*}

\begin{figure*}[h]
    \centering
    \includegraphics[width=1\linewidth]{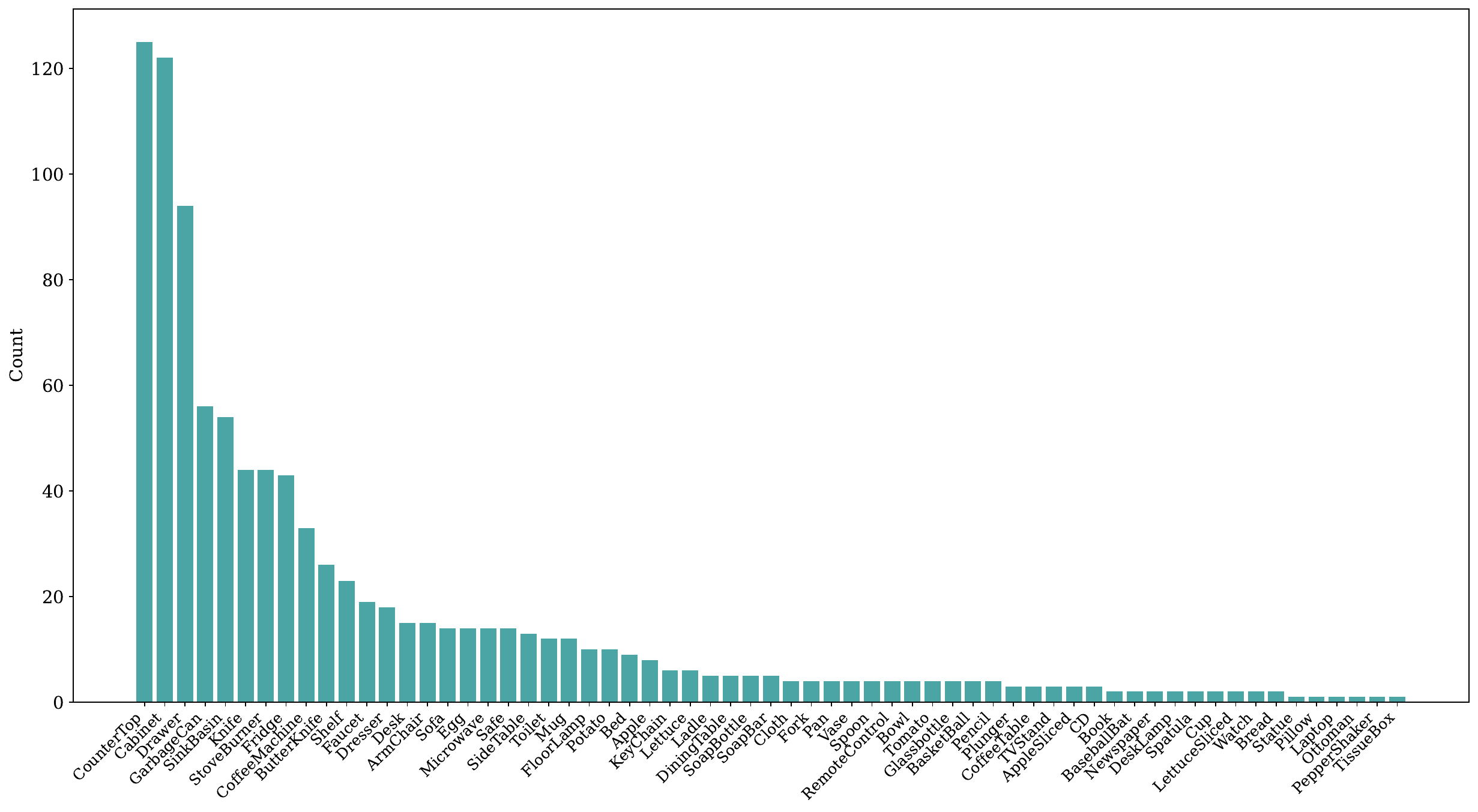}
    \caption{Histogram of objects appearing in all scenes in the dataset}
    \label{fig:obj_count}
\end{figure*}

For the dataset used for action planner evaluation, we plot the histogram of actions and task types in \autoref{fig:action_count}. For \textbf{Action Type} (Left), \texttt{Pickup} and \texttt{Put} are the most frequent actions, as most task types necessitate these actions for object interaction. \texttt{Toggle} and \texttt{OpenClose} are merged from the canonical actions \texttt{ToggleOn}+\texttt{ToggleOff}, and \texttt{OpenObject}+\texttt{CloseObject}, respectively. We note that \texttt{ToggleOff} and \texttt{CloseObject} was always successful for the FILM agent, as these actions are attempted at the same location where the preconditioning action (\texttt{ToggleOn}, \texttt{OpenObject}) was successful. Merging the related actions help balance out the Accept/Reject label distribution per action category.

Similarly in \autoref{fig:obj_count}, we describe object frequencies in the object detector evaluation dataset. Large receptacle objects like \texttt{CounterTop} and \texttt{Cabinet} are observed the most frequently.

\subsection{Action Planner Evaluation}
\label{app:alfred_action}

\autoref{prompt:alfred_action} shows an example prompt used for action planner evaluation. The prompt consists of general task instructions, a docstring explaining how the Planner API works, the agent's status on task progress. For the \textcolor{c2}{Disclaimer} setting, it is informed that the planner can make mistakes. In the \textcolor{c4}{Confidence} setting, a confidence score is 
provided alongside the predicted action, which is the success rate of the past five actions. We additionally note that this confidence score may not always align well with tool success rates in this setting, which might be one reason why the Confidence prompt underperforms the Oblivious prompt in \autoref{tab:alfred_result}. The \textcolor{c1}{Checklist} lists the common failure modes of the planner suggested action. The previous four actions and their success/failure is also presented. Our analysis into the reasoning steps of the LLM shows that models are capable of inferring the robot's state based on this information (e.g., \texttt{[(Open, Fail), (MoveAhead, Success), (Open, Fail), (MoveAhead, Success)] -> Reasoning: ... the previous attempts suggest that the robot might have been trying to open the microwave from too far away}).

\begin{figure}[t]
    \centering
    \includegraphics[width=1\linewidth]{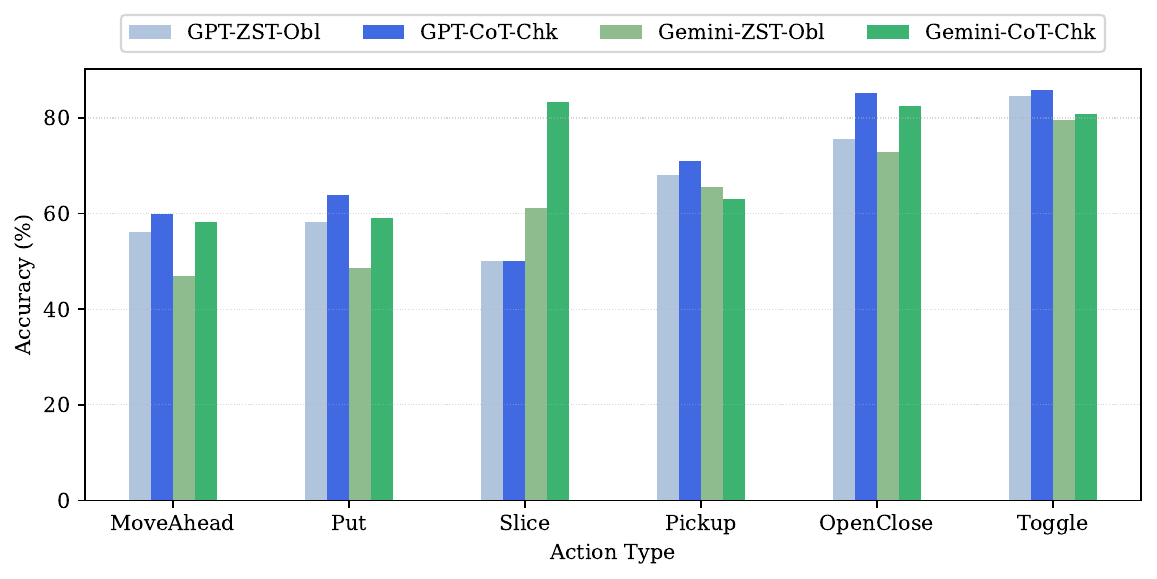}
    \caption{Tool evaluation accuracy on the action planner output binned by action types. We plot the baseline (Zero-shot+Oblivious) with the best performing setting (CoT+Checklist) of the two models.}
    \label{fig:alfred_action}
\end{figure}

In \autoref{fig:alfred_action}, we further analyze the tool evaluation accuracy per different action type. Actions require different preconditions to succeed. 
For instance, successful Pickup, demands target object in the agent's view, within reachable distance, while the agent's hand is empty. 
Thus, different actions require varying levels of spatial reasoning, object/scene detection, and task understanding for assessing feasibility. Compared to interaction actions that may require all the aforementioned capabilities, navigation actions like MoveAhead might be expected as the easiest to infer feasibility, as it mostly relies on spatial reasoning of obstacles. Surprisingly, we find that this is not the case -- because evaluating MoveAhead solely depends on spatial information, it is in fact harder to evaluate compared to other interaction actions, the model having less hints to utilize. For interaction actions, models were able to predict tool success based on objects, which compensates their limited spatial reasoning capability. 

\begin{figure*}[htp]
\centering
\begin{lstlisting}
A robot is working on household tasks in a simulator environment. The robot follows a series of low-level actions to accomplish the task. The robot uses an external tool, a low-level action planner, which predicts the next action to follow. The provided image is a first-person view from the robot's perspective. Refer to the tool suggested action below and decide whether to accept or reject the tool output, based on your judgement of whether the action would succeed/fail.

(*@\textcolor{c2}{The tool can sometimes give incorrect answers. Please cross-check the output based on the image and robot state, to verify the correctness}@*) 
(*@\textcolor{c2}{and feasibility of the planner's output.}@*)
(*@\textcolor{c4}{The tool's prediction confidence (between 0 and 1) is also provided, which may hint the correctness of the output. Confidence is based on previous}@*) 
(*@\textcolor{c4}{action attempts and success/failure.}@*)

(*@\textcolor{c1}{The following are some scenarios where the Planner action might fail.}@*)
(*@\textcolor{c1}{1. Interaction actions might fail if the object is too far from you. In this case, you need to approach closer to the object.}@*)
(*@\textcolor{c1}{2. Interaction actions might fail when you do not have a good view of the object.}@*)
(*@\textcolor{c1}{3. If another object is in your path, MoveAhead will fail due to collision. In this case, you need to walk around the obstacle.}@*)
(*@\textcolor{c1}{4. If a receptacle is occupied with another object, Put will fail.}@*)

# Tool: Planner API
The Planner API provides a function that takes the task_state, observed_state as input and returns the next suggested action. The action is computed based on the agent and target object's location, based on the robot's internal spatial map. 

## Task
possible_actions = ['MoveAhead', 'Open(Receptacle)', 'Close(Receptacle)', 'Pickup(Object)', 'Put(Object, Receptacle)', 'ToggleOn(Object)', 'ToggleOff(Object)', 'Slice(Object)']

## Robot state
task_state = {
    'task_description': "Pick up a pillow and turn a lamp on.",
    'completed_subgoals': [],
    'current_subgoal': "Pickup Pillow",
    'num_steps_taken': 56
}
print(observed_state)
Current room has: Bed, Pillow on a Bed, Cabinet, Drawer, Dresser, GarbageCan, Shelf, SideTable, Sofa, Pillow on a Sofa.
Previous action attempts: [(MoveAhead, Success), (MoveAhead, Success), (MoveAhead, Success), (MoveAhead, Success)]

## Planner output at current step
output = Planner(task_state, observed_state)
print(output)
Pickup(Pillow)(*@\textcolor{c4}{, 0.8}@*)

# Format
Return your answer in this format:
Tool output: [ACTION]
Thought: Your reasoning process
Evaluation: Accept/Reject

The evaluation is a single word indicating whether you accept or reject the tool output. Do not provide any reasoning in the evaluation. Provide your reasoning in the thought section.

# Answer
\end{lstlisting}
\caption{\textbf{Example Prompt for Planner Error Detection} The model is provided instructions to evaluate the output of the Planner and decide whether to Accept or Reject. We denote the instructions specific to the different types of in-context interventions as \textcolor{c2}{Disclaimer}, \textcolor{c4}{Confidence}, and \textcolor{c1}{Checklist}).}
\label{prompt:alfred_action}
\end{figure*}

\subsection{Object Detector Evaluation}
\label{app:alfred_perception}

\autoref{prompt:alfred_perception} shows an example prompt used for object detector evaluation. Similar to the action planner prompt in \autoref{prompt:alfred_action}, general instructions, tool docstring, robot states are given. The robot state here additionally includes the remaining subgoals, as it is helpful in determining which objects are task relevant or not. For instance, while the current subgoal is \texttt{('Pickup', 'Apple')}, correcting detection mistakes for \texttt{'Microwave'} would be beneficial, as it is needed in future subgoals. For Oblivious, \textcolor{c2}{Disclaimer}, and \textcolor{c1}{Checklist}, the tool output is given in a nested dictionary format, where objects are binned into \texttt{'detected'} and \texttt{'filtered'}, based on the detector threshold. For the \textcolor{c4}{Confidence} setting, the detection results are provided in a single dictionary, with objects and their respective raw confidence scores. The instruction mentions that objects with score below 60 will be filtered out. Based on the raw scores, the LLM has to interpret whether specific objects will be kept or discarded.

\begin{figure}[t]
    \centering
    \includegraphics[width=1\linewidth]{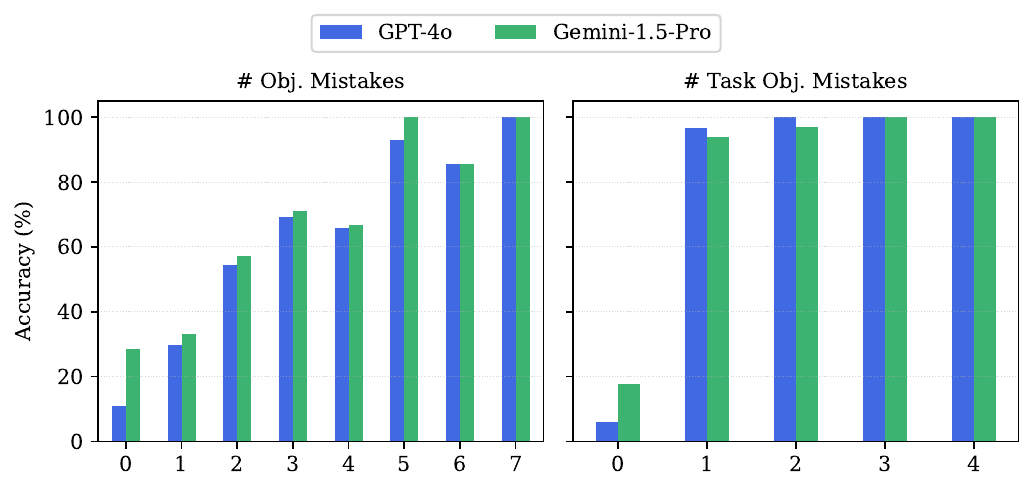}
    \caption{Tool evaluation accuracy on the object detector output binned by the number of detector mistakes on all objects (Left) and task-relevant objects (Right)}
    \label{fig:alfred_perception}
\end{figure}

In \autoref{fig:alfred_perception}, we plot the LLM's evaluation accuracy with respect to the number of mistakes made by the detector, which is one indication of the deviation, $d(o, o^*)$. As the number of detection mistakes increase, it is indeed easier for models to evaluate tool correctness. However, we find that models tend to reject even many acceptable tool outputs where the mistake is not crucial, with the accuracy being extremely low when the number of mistakes are zero in both plots. The models seem to understand when the tool is wrong, but struggles with telling apart task-critical vs tolerable tool mistakes.

\begin{figure*}[htp]
\centering
\begin{lstlisting}
A robot is working on household tasks in a simulator environment. The provided image is a first-person view from the robot's perspective. The robot uses an external tool, an object detector to identify which objects are in the current scene. Refer to the tool output below and evaluate the correctness of the detector with respect to the provided image, and decide whether to accept or reject the tool output. If objects important to the task are ignored by the detector, the tool output should be rejected. Mistakes with regard to task-irrelevant mistakes are acceptable.

(*@\textcolor{c2}{The tool can sometimes give incorrect answers. Please cross-check the output based on the image and robot state, to verify the correctness of the}@*) 
(*@\textcolor{c2}{detector's output.}@*)
(*@\textcolor{c4}{The tool's prediction confidence (between 0 and 100) is also provided, which may hint the correctness of the output. Keep in mind that objects with }@*)
(*@\textcolor{c4}{confidence scores below 60 will be ignored.}@*)

(*@\textcolor{c1}{The following are common examples where the detector mistakes may hinder the robot's ability to accomplish the task. Consider these cases in your }@*)
(*@\textcolor{c1}{reasoning steps.}@*)
(*@\textcolor{c1}{1. Missing task-relevant objects in the scene. In particular, small objects (e.g., keys, credit card) are prone to be missed.}@*)
(*@\textcolor{c1}{2. Hallucinating task-relevant objects that are not in the scene. For example, objects that are similar in shape or color (e.g., apple vs tomato) may}@*) 
(*@\textcolor{c1}{be mistaken.}@*)

# Tool: Object Detector API
The Detector API provides a function that takes the current_image as input and returns the list of objects detected in the image. The obj_categories and receptacles are predefined as below. The prediction consists of two parts: the predicted objects and the filtered objects. The 'filtered' objects are object detections ignored as the detection confidence was lower than the threshold. Only the 'detected' objects will be passed on.

Detector.obj_categories = ['AlarmClock', 'Apple', 'AppleSliced', 'BaseballBat', 'BasketBall', 'Book', 'Bowl', 'Box', 'Bread', 'BreadSliced', 'ButterKnife', 'CD', 'Candle', 'CellPhone', ... ]
Detector.receptacles = ['ArmChair', 'BathtubBasin', 'Bed', 'Cabinet', 'Cart', 'CoffeeMachine', 'CoffeeTable', 'CounterTop', 'Desk', 'DiningTable', 'Drawer', 'Dresser', 'Fridge', ... ]

## Robot state
task_state = {
    'task_description': "Place a cooked apple into the sink.",
    'completed_subgoals': [('Pickup', 'Apple')],
    'remaining_subgoals': [('Open', 'Microwave'), ('Put', 'Microwave'), ('Close', 'Microwave'), ('ToggleOn', 'Microwave'), ('ToggleOff', 'Microwave'), ('Open', 'Microwave'), ('Pickup', 'Apple'), ('Close', 'Microwave'), ('Put', 'SinkBasin')],
    'num_steps_taken': 235
}

## Detector output on current image
Detector(current_image)
# (*@\textcolor{c4}{\texttt{\{'Apple': 3.09, 'Knife': 0.55, 'CounterTop': 63.31, 'DiningTable': 47.09}\}} for \textcolor{c4}{Confidence}@*)
# other prompting methods:
{
    'detected': {'CounterTop'},
    'filtered': {'DiningTable', 'Apple', 'Knife'}
} 

# Format
Return your answer in this format:
Thought: Your reasoning process on the provided information (image, task_state and tool_output)
Evaluation: Accept/Reject

The evaluation is a single word indicating whether you accept or reject the tool output. Do not provide any reasoning in the evaluation. Provide your reasoning in the thought section.

# Answer
\end{lstlisting}
\caption{\textbf{Example Prompt for Object Detector Error Detection} The model is provided instructions to evaluate the output of the Object Detector and decide whether to Accept or Reject. We denote the instructions specific to the different types of in-context interventions as \textcolor{c2}{Disclaimer}, \textcolor{c4}{Confidence}, and \textcolor{c1}{Checklist}. 
}
\label{prompt:alfred_perception}
\end{figure*}

\end{document}